\documentclass{ieeeaccess}
\usepackage{cite}
\usepackage{color}
\usepackage{bbding}
\usepackage{amsmath}
\usepackage{amssymb}
\usepackage{amsfonts}
\usepackage[noend]{algpseudocode}
\usepackage{algorithmicx,algorithm}
\usepackage{comment}
\usepackage[pdftex,linkcolor=blue,citecolor=blue,backref=None]{hyperref}
\usepackage{enumerate}
\usepackage{mathtools} 
\usepackage{multirow}
\usepackage{graphicx}
\usepackage{textcomp}

\def\BibTeX{{\rm B\kern-.05em{\sc i\kern-.025em b}\kern-.08em
    T\kern-.1667em\lower.7ex\hbox{E}\kern-.125emX}}
\begin{document}
\history{Date of publication xxxx 00, 0000, date of current version xxxx 00, 0000.}
\doi{10.1109/ACCESS.2022.0122113}

\title{Attention-based Graph Convolution Fusing Latent Structures and Multiple Features for Graph Neural Networks}
\author{\uppercase{Yang Li}\authorrefmark{1} and \uppercase{Yuichi Tanaka \authorrefmark{2,3}} \IEEEmembership{Senior Member, IEEE}}

\address[1]{Tokyo University of Agriculture and Technology, Tokyo, Japan}
\address[2]{Osaka University, Osaka, Japan}
\address[3]{PRESTO, Japan Science and Technology Agency, Saitama, Japan}
\tfootnote{This work was supported in part by the Japan Science and Technology Agency (JST) PRESTO under Grant JPMJPR1935, and in part by Japan Society for the Promotion of Science (JSPS) KAKENHI under Grant 20H02145. This work was also supported in part by the Japan Science and Technology Agency (JST) FLOuRISH.}

\markboth
{Author \headeretal: Preparation of Papers for IEEE TRANSACTIONS and JOURNALS}
{Author \headeretal: Preparation of Papers for IEEE TRANSACTIONS and JOURNALS}

\corresp{Corresponding author: Yang Li (e-mail: lytuat@msp-lab.org).}

\begin{abstract}
We present an attention-based spatial graph convolution (AGC) for graph neural networks (GNNs). Existing AGCs focus on only using node-wise features and utilizing one type of attention function when calculating attention weights. Instead, we propose two methods to improve the representational power of AGCs by utilizing 1) structural information in a high-dimensional space and 2) multiple attention functions when calculating their weights. The first method computes a local structure representation of a graph in a high-dimensional space. The second method utilizes multiple attention functions simultaneously in one AGC. Both approaches can be combined. We also propose a GNN for the classification of point clouds and that for the prediction of point labels in a point cloud based on the proposed AGC. According to experiments, the proposed GNNs perform better than existing methods. Our codes open at \url{https://github.com/liyang-tuat/SFAGC}.
\end{abstract}


\titlepgskip=-21pt

\maketitle

\section{Introduction}
\label{sec:introduction}
\PARstart{W}{e} often encounter irregularly structured data (signals) in the real world where they do not have a fixed spatial sampling frequency. Such data include opinions on social networks, the number of passengers on traffic networks, coordinates of 3D point clouds, and so on.

\begin{figure}[t]
\setlength{\abovecaptionskip}{0pt}
\setlength{\belowcaptionskip}{0pt}
\centering
\includegraphics[width=7cm]{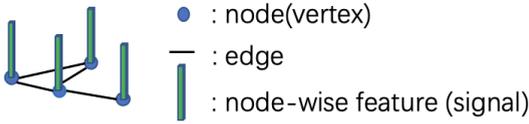}
\caption{An example of graph-structured data.}
\label{fig:graph data}
\end{figure}

Deep neural networks have been widely used in recent years to detect, segment, and recognize regular structured data \cite{Burna2013, video-cnn, YOLO}. However, classical deep learning methods can not directly process the irregularly structured data mentioned above. They can be mathematically represented as data associated with a \textit{graph}. An example of graph-structured data, that is, \textit{graph signals}, is shown in Fig. \ref{fig:graph data}. Deep neural networks for graph signals are called graph neural networks (GNNs): They have received a lot of attention \cite{gnnssurvey, graphsurvey, representation, relational}.

GNNs typically contain multiple graph convolution (GC) layers. The primary mechanism of GCs is to iteratively aggregate (i.e., filter) features from neighbors before integrating the aggregated information with that of the target node \cite{gnnssurvey, graphsurvey, shuman2013emerging, cheung2018graph}. In many existing GC methods, the node-wise features are typically utilized \cite{chebvnet, cayleynets, gcn, graphsage}. Furthermore, it is observed that GCs are a special form of Laplacian smoothing \cite{over-smoothing-deeper}.
This low-pass filtering effect often results in over-smoothing \cite{gnnssurvey, graphsurvey,over-smoothing-deeper}. Over-smoothing means that the node-wise feature values become indistinguishable across nodes. 
Intuitively, representational power of GCs refers to the ability to distinguish different nodes \cite{powerful}. Therefore, over-smoothing may negatively affect the performance of GNNs.

To improve the representational power of GCs, attention-based spatial graph convolutions (AGCs) such as graph attention networks (GATs) \cite{gat} have been proposed. AGCs are believed to have high representational power than the direct spatial methods because they can use features on neighboring nodes through the attention weights. However, there exist two major limitations in existing AGCs: 1) They may lose the \textit{structural information} of the surrounding neighboring nodes, especially in a high-dimensional space. 2) When calculating attention weights for each neighboring node, only one type of attention function is used, e.g., dot-production, subtraction, or concatenation. Different types of attention functions will lead to different attention weights, which affect the representational power of AGCs.

In this paper, we propose a new AGC to overcome the above-mentioned limitations. First, we propose a \textit{local structure projection aggregation}. This operation aggregates the structural information of neighboring nodes of a target node. Second, we also propose an AGC that utilizes multiple-type attention functions. We can simultaneously utilize these two methods to present the attention-based graph convolution fusing latent structures and multiple features (SFAGC). Our contributions are summarized as follows:
\begin{enumerate}
 \item 
By using local structure projection aggregation, we can obtain a representation of the local structural information of the graph in high-dimensional space in an AGC. This allows the convolved nodes to contain richer information than existing methods.
\item 
By using multiple-type attention functions simultaneously, we can obtain better attention weights than the single attention function.
\item 
We construct GNNs for graph and node classifications based on the proposed AGC with 3D point clouds.
We demonstrate that GNNs using our method present higher classification accuracies than existing methods through experiments on ModelNet \cite{modelnet} for graph classification and ShapeNet \cite{shapenet} for node classification.
\end{enumerate}

\textit{Notation:}
An undirected and unweighted graph is defined as $\mathcal{G}:=(\mathcal{V},\mathcal{E})$, where $\mathcal{V}$ is a set of nodes, and $\mathcal{E}$ is a set of edges. The adjacency matrix of $\mathcal{G}$ is denoted as $A$. $\widetilde{D}$ is the diagonal degree matrix. The graph Laplacian is defined as $L:=\widetilde{D}-A$. $I_n$ is the matrix whose elements in the diagonal are 1.

Here, $h_v:=[\text{h}_{v1},\dots,\text{h}_{vi},\dots,\text{h}_{vD}]^\text{T} \in \mathbb{R}^D$ represents a feature vector on the node $v\in \mathcal{V}$, and $D$ is the number of features in $h_v$. $co_v:=[\text{co}_{v1},\dots,\text{co}_{vj},\dots,\text{co}_{vC}]^\text{T} \in \mathbb{R}^C$ represents a coordinate of the node $v\in \mathcal{V}$, and $C$ is the dimension of coordinate in $co_v$. 

The non-linearity function is denoted as $\sigma(\cdot)$. The set of neighboring nodes is $\textit{N}(\cdot)$ in which its cardinality is denoted as $|\textit{N}(\cdot)|$. A multilayer perceptron (MLP) layer is represented as $\text{MLP}(\cdot)$. A channel-wise avg-pooling is denoted as $\text{AvgPool}(\cdot)$. A channel-wise max-pooling is denoted as $\text{MaxPool}(\cdot)$. The vector concatenation operation is denoted as $\text{cat}(\cdot)$. The SoftMax operation is represented as $\text{SoftMax}(\cdot)$.

\section{Preliminary}
\label{sec:Preliminary}
In this section, we present related work for the proposed GC.

\subsection{Graph Convolutions}
The mechanism of GCs is based on message passing. Message passing involves iteratively aggregating information from neighboring nodes and then integrating the aggregated information with that of the target node \cite{gnnssurvey,graphsurvey}.

\begin{figure}[t]
\centering
\includegraphics[width=8cm]{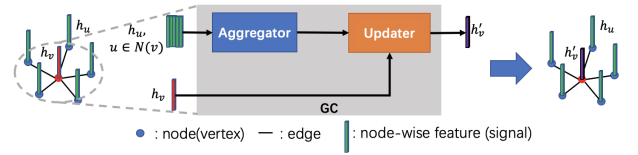}
\caption{A GC has two parts, i.e., an aggregator and an updater. The aggregator is to collects and aggregates the node-wise features of neighboring nodes. The updater merges the aggregated features into the target node to update the node-wise features of the target node.}
\label{fig:message passing}
\end{figure}

Typically, a GC has two parts, i.e., an aggregator and an updater. The aggregator is to collect and aggregate the node-wise features of neighboring nodes. The updater merges the aggregated features into the target node to update the node-wise features of the target node. These two parts are illustrated in Fig. \ref{fig:message passing}. 

Existing GCs can be classified into spectral and spatial methods. Furthermore, spatial methods can be classified into direct and attention-based methods. We briefly introduce them in the following.

\subsubsection{Spectral methods}
In the spectral methods, the aggregation operation is carried out in the graph Fourier domain. Eigenvectors of graph Laplacian are known as the graph Fourier bases \cite{Burna2013, spectralgraphtheory, shuman2013emerging}. In order to reduce the computational complexity for large graphs, polynomial approximations of graph filters are often utilized \cite{chebvnet, waveletschebyv, cayleynets}.

GCN \cite{gcn} further reduces computational complexity through the first-order graph filters. It can be formulated as follows:
\begin{equation}
H_\text{GCN} = \sigma(\widetilde{D}^{-1/2}\widetilde{A}\widetilde{D}^{-1/2}HW),
\end{equation}
where $H := \{\textit{h_v}\}_{\textit{v}\in \mathcal{V}}$ is the set of node-wise features, $\widetilde{A}=A+I_n$ is the adjacency matrix with self-loops, and $W$ is a learnable weight matrix. 

\subsubsection{Spatial methods}
Spatial methods are a counterpart of spectral methods. As we described above, spatial methods can be classified into direct and attention-based methods. Direct spatial methods directly use the node-wise features and the aggregation operation is carried out spatially. A representative method for direct spatial methods is GraphSAGE \cite{graphsage}, which treats each neighbor equally with mean aggregation.

Later, attention-based spatial methods are proposed. Instead of treating all neighboring nodes equally, attention-based methods calculate an attention weight for each neighboring node. Then, they use the weighted sum to aggregate features of neighboring nodes. GAT \cite{gat} is a representative method for attention-based spatial methods. 
It is composed of three steps. In the first step, the learnable weights are multiplied by the node-wise features, i.e., $h^{\prime}_{\mathcal{V}} = \{W \cdot h_v\}_{v \in \mathcal{V}}$, where $W$ is the learnable weights. The second step computes attention weights as follows:
\begin{equation}
a_{uv} = \text{SoftMax}(\sigma({W_a}\cdot(\text{cat}(h^{\prime}_v,h^{\prime}_u)))), u\in N(v)
\end{equation}
where $W_a$ is learnable weights. In the third step, the node-wise features of a target node $v$ is updated as follows:
\begin{equation}
h^{\prime\prime}_v =\sigma\left(\sum_{u\in N(v)}a_{uv}\cdot (h^{\prime}_u)\right).
\end{equation}
However, as mentioned earlier, existing methods ignore the structural information of neighboring nodes in the high-dimensional space, and use only one type of attention function when calculating an attention weight for each neighboring node.

In contrast to the existing methods, we first focus on computing a representation of the structure of neighboring nodes in high-dimensional feature space and installing them into a spatial GC. Second, we use multiple types of attention functions simultaneously in an AGC. We use both of these two methods simultaneously to achieve SFAGC.

\begin{figure}[t]
\setlength{\abovecaptionskip}{0pt}
\setlength{\belowcaptionskip}{0pt}
\centering
\includegraphics[width=8cm]{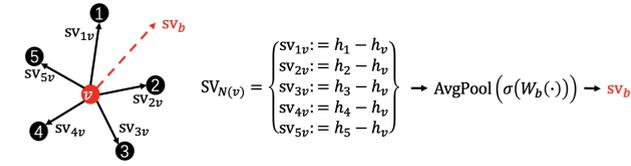}
\caption{An example of a base structure vector. The number in the black circle are the node indices.}
\label{fig:basis vector}
\end{figure}

\begin{figure}[t]
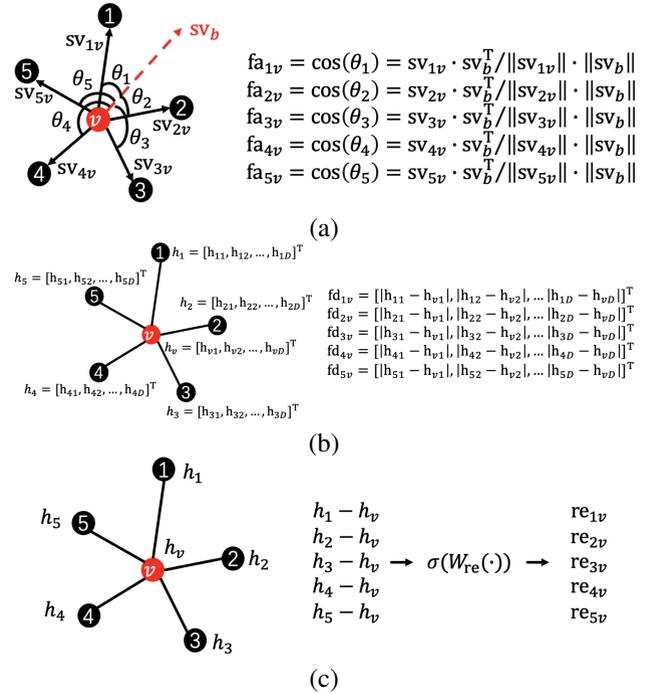

\setlength{\abovecaptionskip}{0pt}
\setlength{\belowcaptionskip}{0pt}
\centering
\begin{minipage}[b]{0.5\textwidth}
\centering
\includegraphics[width=8.2cm]{fa.pdf}\\
(a)
\end{minipage}
\begin{minipage}[b]{0.5\textwidth}
\centering
\includegraphics[width=8.2cm]{fd.pdf}\\
(b)
\end{minipage}
\begin{minipage}[b]{0.5\textwidth}
\centering
\includegraphics[width=7.5cm]{re.pdf}\\
(c)
\end{minipage}
\caption{Example of our structural features. (a) is our feature angle; (b) is our feature distance, $\text{h}_{vi}$ is the element of $h_v$, $D$ is the dimensions of node-wise features; (c) is our relational embedding.}
\label{fig:sf}
\end{figure}

\begin{figure}[t]
\setlength{\abovecaptionskip}{0pt}
\setlength{\belowcaptionskip}{0pt}
\centering
\includegraphics[width=8.3cm]{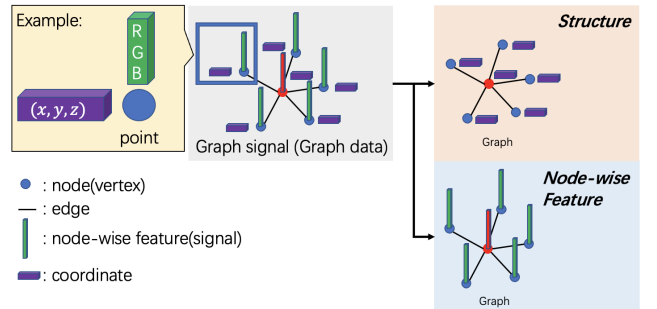}
\caption{An example of a graph in high-dimensional space. A node consists of a coordinate and the node-wise features, i.e., ${\{v:=(co_v,h_v)\}}_{v \in \mathcal{V}}$. For example, the coordinate of a node in a graph of a 3D color point cloud is $(x,y,z)$, and the node-wise features are the values of RGB. We use the coordinates to calculate structural features.}
\label{fig:new graph}
\end{figure}

\begin{figure*}[htb]
\setlength{\abovecaptionskip}{0pt}
\setlength{\belowcaptionskip}{0pt}
\centering
\includegraphics[width=12.5cm]{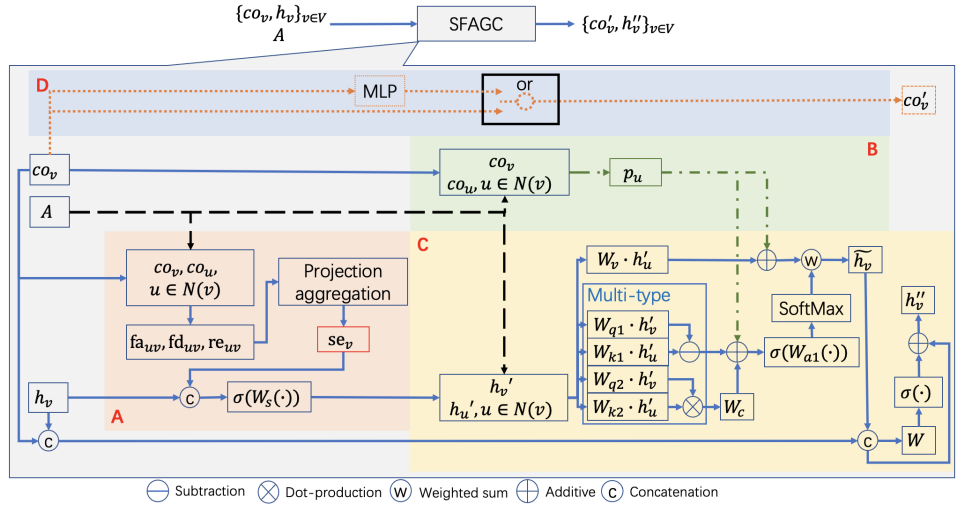}
\caption{The details of our SFAGC.}
\label{fig:SFPMAGC}
\end{figure*}

\begin{figure}[t]
\setlength{\abovecaptionskip}{0pt}
\setlength{\belowcaptionskip}{0pt}
\centering
\includegraphics[width=8cm]{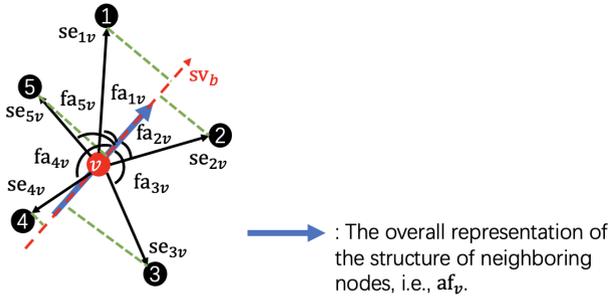}
\caption{An example of the local structure projection aggregation. The $\text{sv}_b$ is the base structure vector defined in (\ref{base vector}).}
\label{fig:projection}
\end{figure}

\subsection{Strctural features}
\label{subsec:Strctural features}
In previous works \cite{SAGC,SAMGC}, three structural features, i.e., feature angle, feature distance, and relational embedding, are proposed to describe the structural information between the target node and neighboring nodes. Below, we briefly introduce them since they are also used in our attention-based GCs.

\subsubsection{Feature angle}
\textit{Feature angle} describes the local structure of neighboring nodes. First, a set of structure vectors pointing from target node $v$ to the neighboring nodes are calculated as $\text{SV}_\textit{N(v)}={\{ \text{sv}_{uv}:=h_u-h_v \}}_{u\in N(v)}$. Then, a base structure vector $\text{sv}_b$ is learned from $\text{SV}_{N(v)}$ as follows:
\begin{equation}
\text{sv}_b=\text{AvgPool}(\{ \sigma(W_b\cdot \text{sv}_{uv})\}_{\text{sv}_{uv}\in \text{SV}_{N(v)}})
\label{base vector}
\end{equation}
where $\textit{W_b}$ is learnable weights. An example of a base structure vector ${sv}_\textit{b}$ is shown in Fig. \ref{fig:basis vector} 

Finally, the cosine of the angle between $\text{sv}_{uv}$ and $\text{sv}_b$ is calculated to obtain the feature angle $\text{fa}_{uv}$ as follows:
\begin{equation}
\text{fa}_{uv}=\cos(\theta_u)=\dfrac{\text{sv}_{uv} \cdot {\text{sv}^{\text{T}}_b}}{{\lVert \text{sv}_{uv}\rVert \cdot \lVert \text{sv}_b \rVert }}, \text{sv}_{uv}\in \text{SV}_{N(v)}
\end{equation}
An example is shown in Fig. \ref{fig:sf} (a).

\subsubsection{Feature distance}
The second structural feature is \textit{feature distance}. It is the absolute difference between the node-wise features of $h_u$ and $h_v$ represented as follows:
\begin{equation}
\text{fd}_{uv}=[|\text{h}_{u1}-\text{h}_{v1}|,...,|\text{h}_{uD}-\text{h}_{vD}|]^\text{T}.
\end{equation}
An example is shown in Fig. \ref{fig:sf} (b).

\subsubsection{Relational embedding}
The third structural feature is \textit{relational embedding}. It can be learned from $\{\text{sv}_{uv}\}$ as follows:
\begin{equation}
\text{re}_{uv}=\sigma(W_\text{re} \cdot \text{sv}_{uv}), u\in N(v).
\end{equation}
where $W_\text{re}$ is learnable weights. An example of it is shown in Fig. \ref{fig:sf} (c).

\begin{algorithm*}[htb]
\small
\caption{SFAGC spatial graph convolution (i.e., forward propagation) algorithm} 
\label{algorithm-SFAGC}
\hspace*{0.02in} {\bf Input:}
graph $\mathcal{G}=(\mathcal{V},\mathcal{E})$; input features $\{h_v, v\in \mathcal{V}\}$;
coordinates $\{co_v, v\in \mathcal{V}\}$;
weight matrices $W_{re}$, $W_b$, $W_{\text{se}}$, $W_s$, $W_{\text{P1}}$, $W_{\text{P2}}$, $W_{q1}$, $W_{k1}$, $W_{q2}$, $W_{k2}$, $W_{a1}$, $W_{a2}$, $W_v$ and $W$; non-linearity function $\sigma(\cdot)$; neighborhood function $N(\cdot)$; MLP layer $\text{MLP}(\cdot)$; channel-wise avg-pooling $\text{AvgPool}(\cdot)$; concatenation operation $\text{cat}(\cdot)$; transposition ${\cdot}^{\text{T}}$; SoftMax operation $\text{SoftMax}(\cdot)$\\
\hspace*{0.02in} {\bf Output:} 
Convolved features $h^{\prime\prime}_v$ and updated coordinates $co^{\prime}_v$ for all $v\in \mathcal{V}$
\begin{algorithmic}[1]
\For{$v\in \mathcal{V}$}
       \State $\text{SV}_{N(v)}:=\{{\text{sv}_{uv}}=co_u-co_v\}$
       \State $\text{sv}_b=\text{AvgPool}(\{\sigma(W_b \cdot (\text{sv}_{uv}))\})$ 
       \State $\text{fa}_{uv}=\cos(\theta_u)=\dfrac{{\text{sv}_{uv}} \cdot {\text{sv}^{\text{T}}_{b}}}{{\lVert {\text{sv}_{uv}}\rVert \cdot \lVert {\text{sv}_{b}} \rVert }}, {\text{sv}_{uv}}\in {\text{SV}_{N(v)}}$ \Comment{feature angle}
       \State $\text{fd}_{uv}=[|\text{co}_{u1}-\text{co}_{v1}|,...,|\text{co}_{uC}-\text{co}_{vC}|]^\text{T}$ \Comment{feature distance}
       \State $\text{re}_{uv}=\sigma(W_{re}\cdot (co_u-co_v)), u\in N(v)$ \Comment{relational embedding}
       \State
       $\text{s}_{uv} =\text{cat}(\text{fd}_{uv},\text{re}_{uv},\sigma(W_{\text{se}} \cdot (\text{cat}(\text{fd}_{uv},\text{re}_{uv}))))$
       \State 
       $\hat{\text{se}}_{uv}= \text{fa}_{uv} \cdot \text{s}_{uv}$
       \State $\text{af}_v = \sum_{u \in N(v)} \hat{\text{se}}_{uv}$ \Comment{local structure projection aggregation}
       \State
       $h^{\prime}_v = \sigma(W_s(\text{cat}(h_v, \text{af}_v)))$ \Comment{structures and node-wise features fusing}
\EndFor
\State {\bf end for}
\For{$v\in \mathcal{V}$}
       \State $p_u = W_{\text{P2}}\cdot(\sigma(W_{\text{P1}}\cdot(co_u - co_v))),u\in N(v)$ \Comment{position embedding}
       \State $\text{qk}_{vu} = W_{q1} \cdot h^{\prime}_v-W_{k1} \cdot h^{\prime}_u+W_c\cdot((W_{q2} \cdot h^{\prime}_v)\cdot {(W_{k2} \cdot h^{\prime}_u)}^{\text{T}})+p_u,u\in N(v)$
       \State $\text{at}_{vu} = \text{SoftMax}(W_{a2}(\sigma(W_{a1}\cdot \text{qk}_{vu})))$ \Comment{attention weights}
       \State $\widetilde{h_v} = \sum_{u\in N(v)} \text{at}_{vu} \cdot (W_{v} \cdot h^{\prime}_u+p_u)$ \Comment{weighted-sum aggregation}
       \State ${h^{\prime\prime}_v}= \sigma(W\cdot\text{cat}(h_v,co_v, \widetilde{h_v}))$ \Comment{node-wise features update}
       \State $co^{\prime}_v=\text{MLP}(co_v)$ \Comment{coordinate update}
\EndFor
\State {\bf end for}
\State $h^{\prime\prime}_v, \forall v\in \mathcal{V}$
\State $co^{\prime}_v, \forall v\in \mathcal{V}$
\State \Return $h^{\prime\prime}_v$ and $co^{\prime}_v$
\end{algorithmic}
\end{algorithm*}

\section{SFAGC}
\label{sec:SFPMAGC}
In this section, we introduce SFAGC. As mentioned above, we have two goals: Utilizing 1) the structural information of neighboring nodes in a high-dimensional feature space during a single step GC and 2) multiple-type attention functions simultaneously when calculating the attention weights. 

Fig. \ref{fig:new graph} illustrates an example of a graph with feature vectors in the spatial domain. Spatially distributed nodes often have their coordinates and associated node-wise features, i.e., ${\{v:=(co_v,h_v)\}}_{v \in \mathcal{V}}$, where $co_v$ is the coordinate of the node $v$. For example, a 3D color point cloud equips a 3-D $(x,y,z)$ coordinate and its node-wise features as RGB values. In the previous structure-aware GCs \cite{SAGC,SAMGC}, the node-wise features are simply used as their coordinates. In contrast, the proposed GC, SFAGC, simultaneously considers the coordinates and node-wise features. 

To achieve our goals, the SFAGC has four parts: 
\begin{description}
\item [\textit{A}.]  Local structure projection aggregation and fusing
\item [\textit{B}.]  Position embedding
\item [\textit{C}.]  Weighted sum aggregation and update
\item [\textit{D}.]  Coordinates processing
\end{description}
SFAGC is illustrated in Fig. \ref{fig:SFPMAGC} and the algorithm of SFAGC is summarized in Algorithm \ref{algorithm-SFAGC}. We sequentially introduce these parts.

\subsection{Local structure projection aggregation and fusing}
\label{subsec:local structure projection aggregation and fusing}
We propose a projection aggregation operation to obtain the structure representation in the feature space. We then fuse this with the node-wise features of the target node. 

\subsubsection{Local structure projection aggregation}
The inputs of this step are the feature angle, feature distance and relational embedding, i.e., $\text{fa}_{uv}$, $\text{fd}_{uv}$ and $\text{re}_{uv}$, introduced in Section \ref{subsec:Strctural features}. We first compute structure vectors as follows:
\begin{equation}
\text{s}_{uv} =\text{cat}(\text{fd}_{uv},\text{re}_{uv},\sigma(\textit{W}_\text{se}(\text{cat}(\text{fd}_{uv},\text{re}_{uv})))),u\in N(v),
\end{equation}
where $W_\text{se}$ is a learnable weight matrix.

Then, we project each $\text{s}_{uv}$ as follows:
\begin{equation}
\hat{\text{se}}_{uv}= \text{fa}_{uv} \cdot \text{s}_{uv},u\in N(v).
\end{equation}

Finally, we calculate the summation of the projected structure vectors as follows:
\begin{equation}
\text{af}_v = \sum_{u \in N(v)} \hat{\text{se}}_{uv}.
\end{equation}
Fig. \ref{fig:projection} illustrates an example of the local structure projection aggregation. 

\subsubsection{Fusing structure information with node-wise features}
In this step, we fuse the $\text{af}_v$ with the $h_v$ as follows:
\begin{equation}
h^{\prime}_v = \sigma(W_s(\text{cat}(h_v, \text{af}_v))), v \in \mathcal{V}.
\end{equation}
where $W_s$ is learnable weights.

\subsection{Position embedding}
\label{subsec:position embedding}
Position encoding is crucial for a self-attention mechanism because it enables the aggregation operation to adapt to local data structure \cite{Transformer,exploring-self-attention}. Our method directly learns a position embedding, while existing methods use cosine function-based position encoding \cite{exploring-self-attention}.

We embed the difference of the coordinates between the neighboring node $u$ and target node $v$ in the feature space. Our position embedding $p_u$ for the node $u$ is represented as follows:
\begin{equation}
p_u = W_\text{P2}\cdot(\sigma(W_\text{P1}\cdot(co_u-co_v))), u \in N(v)
\end{equation}
where $W_\text{P1}$ and $W_\text{P2}$ are learnable weights.

\subsection{Weighted sum aggregation and update}
\label{subsec:Weighted sum aggregation and update}
In this part, we update the node-wise features of the target node $v$.

First, we introduce the calculation steps of the attention weights. Then, we present the weighted sum aggregation and node-wise features update step used in SFAGC.

\subsubsection{Attention weights}
As we described above, we simultaneously use multiple-type attention functions to calculate attention weights. In existing methods \cite{Transformer,exploring-self-attention,PointTransformer}, the subtraction or dot-production attention function is often utilized to calculate attention weights. Instead of the single attention function, we use these attention functions simultaneously. The subtraction attention function is defined as
\begin{equation}
\text{a}_{1vu} := W_{q1} \cdot h^{\prime}_v - W_{k1} \cdot h^{\prime}_u, u \in N(v),
\end{equation} 
where $W_{q1}$ and $W_{k1}$ are learnable weights. The dot-production attention function is represented as
\begin{equation}
\text{a}_{2vu} := (W_{q2} \cdot h^{\prime}_v) \cdot {(W_{k2} \cdot h^{\prime}_u\text)}^{T}, u \in N(v)
\end{equation}
where $W_{q2}$ and $W_{k2}$ are learnable weights.

Then, the two types of attention functions are added with the position embedding $\textit{p_u}$ as follows:
\begin{equation}
\text{qk}_{vu} = \text{a}_{1vu}+W_c \cdot \text{a}_{2vu} + p_u, u \in N(v)
\end{equation}
where $W_c$ is learnable weights that also converts $\text{a}_{2vu}$ into the same dimensions as $\text{a}_{1vu}$.

Finally, $\text{qk}_{vu}$ is input into a small network to calculate the attention weights between the target node $v$ and the neighboring node $u$ as follows:
\begin{equation}
\text{at}_{vu} = \text{SoftMax}\left(\frac{W_{a2}\cdot\sigma(W_{a1}\cdot \text{qk}_{vu})}{\sqrt{d_{out}}}\right), u \in N(v),
\end{equation}
where $W_{a1}\in\mathbb{R}^{d_{\text{in}}\times d_{out}}$ and $W_{a2}\in\mathbb{R}^{d_{out}\times 1}$ are learnable weights, $d_{in}$ and $d_{out}$ are the dimensions of the $W_{a1}$.

\subsubsection{Weighted-sum aggregation}For $h^{\prime}_v$ and $h^{\prime}_u,u\in N(v)$, weighted sum aggregation is calculated as follows:
\begin{equation}
\widetilde{h_v} = \sum_{u \in N(v)}\text{at}_{vu} \cdot (W_v \cdot h^{\prime}_u+p_u),
\end{equation}
where $W_v$ is a learnable matrix.

\subsubsection{Node-wise features update}
$h_v$, $co_v$ and $\widetilde{h_v}$ are integrated as follows:
\begin{equation}
h^{\prime\prime}_v = \sigma(W\cdot \text{cat}(h_v,co_v,\widetilde{h_v})).
\end{equation}
where $W$ is learnable weights.

\subsection{Coordinate update}
\label{subsec:Coordinates updating}
Finally, we update the coordinate of the target node $v$ as follows:
\begin{equation}
co^{\prime}_v=\text{MLP}(co_v), v\in \mathcal{V}.
\end{equation}

Hereafter, we represent the set of these operations as $\{{h^{\prime\prime}_\mathcal{V},co^{\prime}_\mathcal{V}}\} :=\text{SFAGC}(h_\mathcal{V}, co_\mathcal{V}, A)$.

\begin{figure*}[htb]
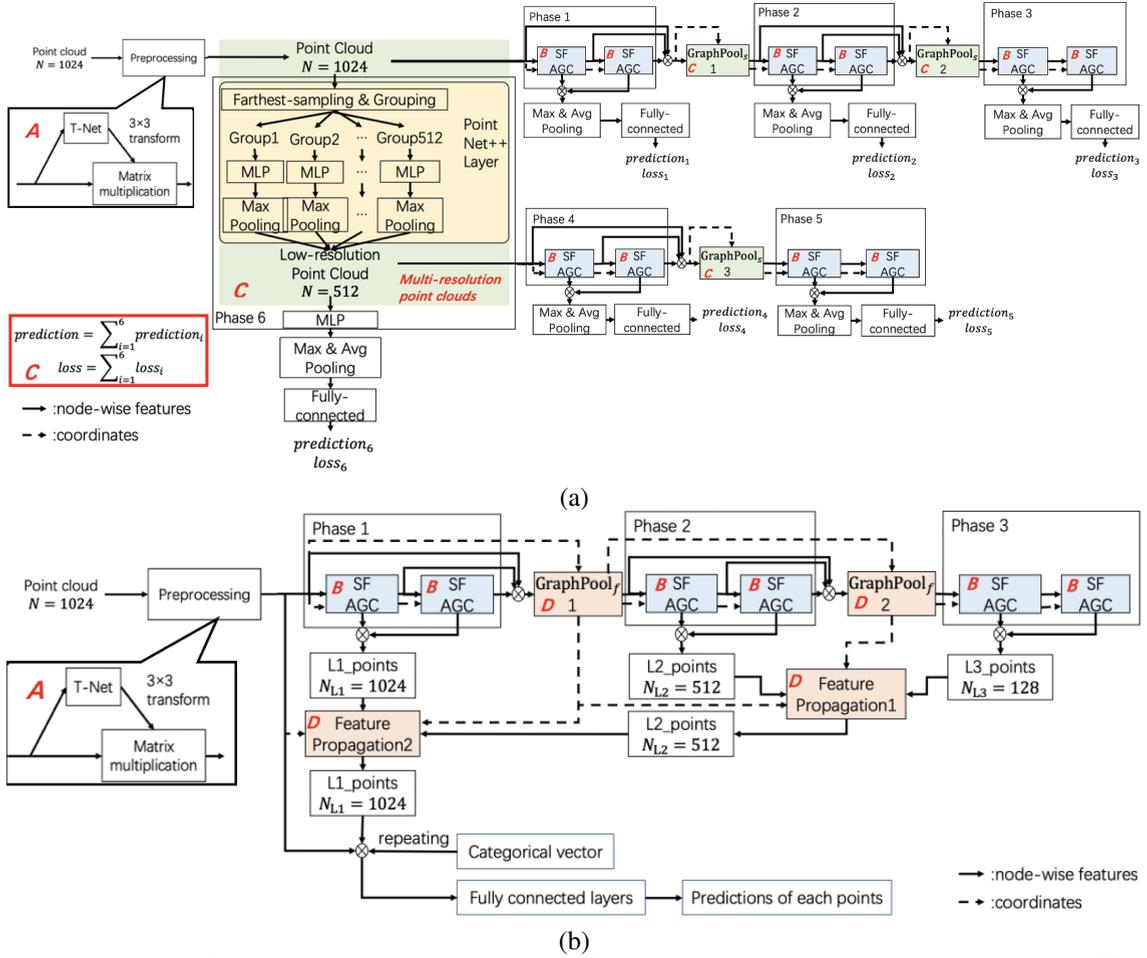

\setlength{\abovecaptionskip}{0pt}
\setlength{\belowcaptionskip}{0pt}
\centering
\begin{minipage}[b]{1.0\textwidth}
\centering
\includegraphics[width=15cm]{3DC-SFPMAGC.pdf}\\
(a)
\end{minipage}
\begin{minipage}[b]{1.0\textwidth}
\centering
\includegraphics[width=15cm]{3DP-SFPMAGC.pdf}\\
(b)
\end{minipage}
\caption{The architectures of our 3D point cloud classification network and our 3D point cloud segmentation network. (a) is the architecture of 3D point cloud classification network; (b) is the architecture of 3D point cloud segmentation network, $\text{L}p\_\text{points}$ is the set of outputs of the $p$th phase. $N_{\text{L}p}$ is the number of nodes of the $\text{L}p\_\text{points}$.}
\label{fig:networks}
\end{figure*}

\begin{figure}[t]
\setlength{\abovecaptionskip}{0pt}
\setlength{\belowcaptionskip}{0pt}
\centering
\includegraphics[width=8.3cm]{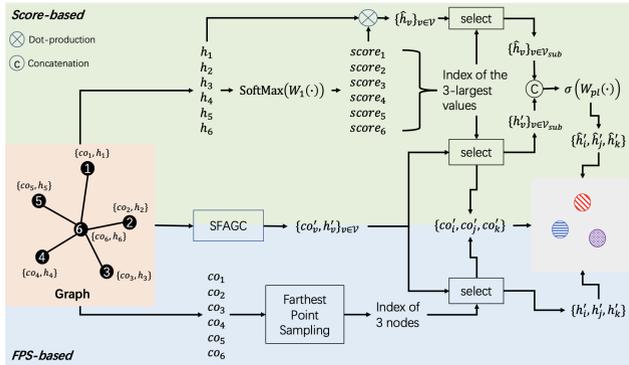}
\caption{The details of our graph pooling operation. The score-based graph pooling is used in 3D point cloud classification network. The FPS-based graph pooling is used in 3D point cloud segmentation network. Here, we set $t=3$.}
\label{fig:graph-pooling}
\end{figure}

\begin{table*}[htb]
\centering  
\footnotesize
\setlength{\abovecaptionskip}{0pt} 
\setlength{\belowcaptionskip}{10pt}
\caption{The hyperparameters of the point cloud classification network. $k$ is the value of $k$-NN, $t$ is the number of selected nodes. For Pointnet++ layer, $S$ is the number of sampled points, $r$ is the radius of each group, $D$ is the number of points of each group. $\text{co}_{\text{in}}$ is the number of channels of the input coordinates. $\text{f}_{\text{in}}$ is the number of channels of the input node-wise features. $\text{co}_{\text{out}}$ is the number of channels of the output coordinates. $\text{f}_{\text{out}}$ is the number of channels of the output node-wise features. The symbol ‘‘-’’ indicates that the parameters are not available. The input 3D point cloud is $\mathcal{X}=\{\text{x}_i\}_{i=1}^N$.} 
\label{3dc-hp}
\setlength{\tabcolsep}{2.7mm}{
 \begin{tabular}{r|l|l|l|l|l}
     \hline
     \hline
       Epoch & \multicolumn{5}{|l}{200} \\
       \hline
       Batch size & \multicolumn{5}{|l}{16} \\
       \hline
       Learning rate & \multicolumn{5}{|l}{0.001} \\
       \hline
       Drop out & \multicolumn{5}{|l}{0.3} \\
       \hline
        Phase & Graph convolution layer & Coordinates update & $\textit{k}$ & \multicolumn{2}{|l}{[$\text{co}_\text{in}$,$\text{f}_\text{in}$,$\text{co}_\text{out}$,$\text{f}_\text{out}$]}    \\
       \hline
        \multirow{2}*{Phase1} & SFAGC($co_v=x_v$) & $co^{\prime}_v=\text{MLP}(co_v)$ & 20 & \multicolumn{2}{|l}{[3,3,32,64]} \\
         & SFAGC & $co^{\prime}_v=co_v$ & 20 & \multicolumn{2}{|l}{[32,64,-,64]} \\
       \hline
        \multirow{2}*{Phase2} & SFAGC & $co^{\prime}_v=\text{MLP}(co_v)$ & 20 & \multicolumn{2}{|l}{[32,64,64,64]} \\
         & SFAGC & $co^{\prime}_v=co_v$ & 20 & \multicolumn{2}{|l}{[64,64,-,128]} \\
       \hline
        \multirow{2}*{Phase3} & SFAGC & $co^{\prime}_v=\text{MLP}(co_v)$ & 20 & \multicolumn{2}{|l}{[64,128,128,256]} \\
         & SFAGC & $co^{\prime}_v=co_v$ & 20 & \multicolumn{2}{|l}{[128,256,-,256]} \\
       \hline
        \multirow{2}*{Phase4} & SFAGC & $co^{\prime}_v=\text{MLP}(co_v)$ & 20 & \multicolumn{2}{|l}{[64,128,64,128]} \\
         & SFAGC & $co^{\prime}_v=co_v$ & 20 & \multicolumn{2}{|l}{[64,128,-,128]} \\
       \hline
        \multirow{2}*{Phase5} & SFAGC & $co^{\prime}_v=\text{MLP}(co_v)$ & 20 & \multicolumn{2}{|l}{[128,128,128,256]} \\
         & SFAGC & $co^{\prime}_v=co_v$ & 20 & \multicolumn{2}{|l}{[128,256,-,256]} \\
        \hline
        Graph pooling layername & Graph convolution layer & Coordinates update & $\textit{k}$ & $\textit{t}$ & [$\text{co}_\text{in}$,$\text{f}_\text{in}$,$\text{co}_\text{out}$,$\text{f}_\text{out}$]    \\
       \hline
        $\text{GraphPool}_s$ 1 & SFAGC($co_v=h_v$) & $co^{\prime}_v=\text{MLP}(co_v)$ & 36 & 512 &[131,131,32,64] \\
        $\text{GraphPool}_s$ 2 & SFAGC($co_v=h_v$) & $co^{\prime}_v=\text{MLP}(co_v)$ & 64 & 128 &[320,320,64,128] \\
        $\text{GraphPool}_s$ 3 & SFAGC($co_v=h_v$) & $co^{\prime}_v=\text{MLP}(co_v)$ & 64 & 128 &[384,384,128,128] \\
       \hline
        Pointnet++ layer name & $\textit{S}$ & $\textit{r}$ & $\textit{D}$ & \multicolumn{2}{|l}{[input channels, output channels]} \\
        \hline
        \multirow{2}*{Pointnet++} & \multirow{2}*{512} & 0.2 & 32 & \multicolumn{2}{|l}{[3,64]} \\
         & & 0.4 & 128 & \multicolumn{2}{|l}{[3,64]} \\
        \hline
        \hline
   \end{tabular}}
\end{table*}

\section{Implementation}
\label{sec:Implementation}
In this section, we construct classification and segmentation networks for 3D point clouds based on SFAGC. Their architectures are illustrated in Fig. \ref{fig:networks}.
In the following, first, we introduce the components shared by the two GNNs. Then, specific implementations for each of the GNNs are introduced.

Here, we suppose that the input point cloud is given by $\mathcal{X}=\{x_i\}_{i=1}^N$ where $N$ is the number of points.

\subsection{Preprocessing}
\label{subsec:Preprocessing}
To alleviate effects on rotations of point clouds, we use the same transformation module as PointNet \cite{pn} for preprocessing.

\subsection{SFAGC module}
\label{subsec:SFPMAGC module}
The inputs to this module are a set of features obtained from the previous layer and a set of node coordinates in the feature space. Suppose that the set of features output from the previous layer is ${\mathcal{H}=\{h_i\}_{i=1}^M}$ where $M$ is the number of features of the input. For the first module, $\mathcal{H}=CO=\mathcal{X}$ where ${CO=\{co_j\}_{j=1}^M}$ is the set of node coordinates. 

First, we construct $\mathcal{G}$ with the $k$-NN graph from $CO$ where the $i$th node $v_i$ in $\mathcal{G}$ corresponds to the $i$th feature $h_i$ and the $i$th coordinate $co_i$.

Recent studies \cite{dgcnn,ecc} have shown that dynamic graph convolution, i.e., allowing the graph structure to change at each layer, can perform better than that with a fixed graph structure. Therefore, we update coordinates of nodes at each SFAGC module, and we construct different graphs for different SFAGC modules.

\subsection{Design of 3D point cloud classification network}
\label{subsec:3D point cloud classification network}
Fig. \ref{fig:networks} (a) illustrates the architecture of 3D point cloud classification network based on SFAGC. In the following, we describe the details of the building blocks that are specifically designed for point cloud classification.

\subsubsection{Multi-resolution point clouds}
 We use a layer of PointNet++ \cite{pn++} to generate a low-resolution point cloud. Both global and local information of the point cloud can be obtained through the multi-resolution structure.

\subsubsection{Score-based graph pooling}
\label{subsubsec:score-based graph pooling}
Effective graph pooling methods are a hot topic in GNNs and graph signal processing \cite{graphpoolsurvey,tanaka}. Early work has been done by global pooling of all node-wise features or by using graph coarsening algorithms. 

Recently, trainable graph pooling operations DiffPool \cite{diffpool}, GraphU-net \cite{graphunet}, and AttPool \cite{attpool} have been proposed. Inspired by the ResNeXt \cite{resnext}, we extend the score-based graph pooling module proposed in SAMGC \cite{SAMGC} by introducing the multi-branch architecture. The score-based graph pooling module has three branches: score-based sampling, integration, and SFAGC branches. The architecture of the score-based graph pooling is shown in Fig. \ref{fig:graph-pooling}. In the following, we introduce their details.

In the score-based sampling branch, we propose a scroe-based sampling to find the indices of the best $t$ nodes according to their scores. The score associated with each node is first computed as follows:
\begin{equation}
score_v=\text{SoftMax}(W_1\cdot h_v),v\in\mathcal{V},
\end{equation}
where $W_1$ is learnable weights. We then sort the nodes in descending order according to their scores. We find the indices of the top $t$ nodes as follows:
\begin{equation}
\text{idx}_{select}=\text{rank}(\{score_v\}_\mathcal{V},t),
\end{equation}
where $\text{rank}(\cdot)$ is the ranking operation, and it finds the indices of the $t$ highest scores.

In the integration branch, node-wise features are multiplied by the node scores as follows: 
\begin{equation}
\hat{h}_{\mathcal{V}}=\{{score}_v\cdot h_v\}_{v \in \mathcal{V}}.
\end{equation}

In the SFAGC branch, the input graph is processed using SFAGC as follows:
\begin{equation}
\{h^{\prime}_{\mathcal{V}},{co}^{\prime}_{\mathcal{V}}\}=\text{SFAGC}(h_{\mathcal{V}},co_{\mathcal{V}}).
\end{equation}

Finally, the subset of $\hat{h}_{\mathcal{V}}$ and the subset of $\{h^{\prime}_{\mathcal{V}},{co}^{\prime}_{\mathcal{V}}\}$ are found using $\text{idx}_{select}$ as follows:
\begin{equation}
\begin{aligned}
\hat{h}_{\mathcal{V}_{sub}} &= \hat{h}_{\mathcal{V}}[\text{idx}_{select}] \\
\{h^{\prime}_{\mathcal{V}_{sub}},{co}^{\prime}_{\mathcal{V}_{sub}}\} &= \{h^{\prime}_{\mathcal{V}},{co}^{\prime}_{\mathcal{V}}\}[\text{idx}_{select}]
\end{aligned}
\end{equation}
$\hat{h}_{\mathcal{V}_{sub}}$ and $h^{\prime}_{\mathcal{V}_{sub}}$ are merged using learnable weights as follows:
\begin{equation}
\hat{h}^{\prime}_{\mathcal{V}_{sub}} = \sigma(W_{pl} \cdot \text{cat}(\hat{h}_{\mathcal{V}_{sub}},h^{\prime}_{\mathcal{V}_{sub}})),
\end{equation}
where $W_{pl}$ is learnable weights. The score-based graph pooling can be summarized as follows:
\begin{equation}
\{{co}^{\prime}_{\mathcal{V}_{sub}},\hat{h}^{\prime}_{\mathcal{V}_{sub}}\} = \text{GraphPool}_s(co_\mathcal{V},h_\mathcal{V}).
\end{equation}

\subsubsection{Hierarchical prediction architecture}
Here, we also use the intermediate supervision technique \cite{intermediate-supervision} and the hierarchical prediction architecture (Fig. \ref{fig:networks} (a)), as in SAMGC \cite{SAMGC}. The advantage of this architecture is that, by combining the outcomes of the different phases, more reliable and robust predictions can be produced \cite{SAMGC}. The details are presented below.

We use two SFAGC modules in phase 1 to phase 5. One PointNet++ \cite{pn++} layer is used in phase 6. Each phase connects with a max pooling layer and an average pooling layer. The outputs are then concatenated and input into a fully connected layer. We calculate the prediction and the classification loss for each phase. The total classification loss is obtained by adding the losses of several phases. Meanwhile, the total prediction is also increased by the predictions of several phases. The following is a representation of this processing:
\begin{align}
prediction &= \sum_{i=1}^{P}prediction_i, \\
loss &= \sum_{i=1}^{P}loss_i,
\end{align}
where $prediction_i$ is the prediction of the $i$th phase, $loss_i$ is the cross-entropy loss of the $i$th phase. $prediction$ and $loss$ are the total prediction and classification loss, respectively. $P$ is the number of phases.

\subsection{Design of 3D point cloud segmentation network}
\label{subsec:3D point cloud part segmentation network}
Fig. \ref{fig:networks} (b) illustrates the architecture of the 3D point cloud segmentation network based on SFAGC. In the following, we describe the details of the building blocks that are specifically designed for point cloud segmentation.

\subsubsection{Farthest point sampling-based graph pooling}
\label{subsubsec:FPS-based graph pooling}
For point cloud segmentation, we use the graph pooling module to reduce the overall computation cost. Here, we propose the farthest point sampling-based graph pooling (FPS-based graph pooling) by modifying the score-based graph pooling. The architecture of the FPS-based graph pooling is illustrated in Fig. \ref{fig:graph-pooling}. In the following, we introduce its details.

The FPS-based graph pooling has a multi-branch architecture like the score-based graph pooling in Section \ref{subsubsec:score-based graph pooling}. In contrast to the score-based method, the FPS-based graph pooling has two branches, i.e., the FPS and SFAGC branches.

In the FPS branch, we perform FPS on nodes to obtain indices of the best $t$ nodes according to their coordinates. FPS algorithm is widely utilized in 3D point cloud processing \cite{pn++,pnasnl}. The mechanism of FPS is to iteratively select the node that is farthest from the existing sampled nodes. Therefore, the sampled nodes with the FPS-based sampling are expected to be more evenly distributed than those with the score-based sampling. This branch can be summarized as follows:
\begin{equation}
\text{idx}_{select}=\text{FPS}(\{co_v\}_{\mathcal{V}},t).
\end{equation}
where $\text{idx}_{select}$ is the indices of the $t$ selected nodes, $\text{FPS}(\cdot)$ is the farthest point sampling algorithm.

The SFAGC branch is the same as the one that is used in the score-based graph pooling represented as
\begin{equation}
\{h^{\prime}_{\mathcal{V}},{co}^{\prime}_{\mathcal{V}}\}=\text{SFAGC}(h_{\mathcal{V}},co_{\mathcal{V}}).
\end{equation}

Finally, the subset of $\{h^{\prime}_{\mathcal{V}},{co}^{\prime}_{\mathcal{V}}\}$ are extracted using $\text{idx}_{select}$ as follows:
\begin{equation}
\{co^{\prime}_{\mathcal{V}_{sub}},h^{\prime}_{\mathcal{V}_{sub}}\} = \{co_{\mathcal{V}},h_{\mathcal{V}}\}[\text{idx}_{select}].
\end{equation}
The FPS-based graph pooling is represented as follows:
\begin{equation}
\{co^{\prime}_{\mathcal{V}_{sub}},h^{\prime}_{\mathcal{V}_{sub}}\} = \text{GraphPool}_f(co_\mathcal{V},h_\mathcal{V}).
\end{equation}

\subsubsection{Upsampling operation}
Graph pooling can be regarded as a downsampling operation. For point cloud segmentation, the network also needs to perform upsampling in order to maitain the number of points. Therefore, the feature propagation used in PointNet++ \cite{pn++} is also used in our network as the upsampling operation.

\begin{table*}[htb]  
\footnotesize
\centering  
\setlength{\abovecaptionskip}{0pt} 
\setlength{\belowcaptionskip}{10pt}
\caption{Comparison results of the 3D shape classification on the ModelNet benchmark. OA indicates the average accuracy of all test instances, and mAcc indicates the average accuracy of all shape categories. The symbol ‘‘-’’ indicates that the results are not available from the references.}  
\label{3dc-results}
 \setlength{\tabcolsep}{3.8mm}{
 \begin{tabular}{c|l|l|ll|ll} 
     \hline
     \hline
       Type & \multicolumn{2}{|l|}{Method}  & ModelNet40 &  & ModelNet10 & \\
      \cline{4-7}
           & \multicolumn{2}{|l|}{ } & OA & mAcc & OA & mAcc \\
       \hline
        Pointwise MLP& \multicolumn{2}{|l|}{PointNet \cite{pn}} & 89.2\%\ & 86.2\%\ & - & - \\
        Methods& \multicolumn{2}{|l|}{PointNet++ \cite{pn++}} & 90.7\%\ & - & - & - \\
        & \multicolumn{2}{|l|}{SRN-PointNet++ \cite{srn-pn++}} & 91.5\%\ & - & - & - \\
        \hline
        Transformer-based& \multicolumn{2}{|l|}
        {PointASNL \cite{pnasnl}} & 93.2\%\ & - & 95.9\%\ & - \\
        Methods& \multicolumn{2}{|l|}{PCT \cite{PCT}} &  93.2\%\ & - & - & - \\
        & \multicolumn{2}{|l|}{PointTransformer \cite{PointTransformer}} &  93.7\%\ & 90.6 \%\ & - & - \\
       \hline
        Convolution-based& \multicolumn{2}{|l|}{PointConv \cite{pnconv}} & 92.5\%\ & - & - & - \\
        Methods& \multicolumn{2}{|l|}{A-CNN \cite{a-cnn}} & 92.6\%\ & 90.3\%\ & 95.5\%\ & 95.3\%\ \\
        & \multicolumn{2}{|l|}{SFCNN \cite{sfcnn}} & 92.3\%\ & - & - & - \\
        & \multicolumn{2}{|l|}{InterpCNN \cite{interpcnn}} & 93.0\%\ & - & - & - \\
        & \multicolumn{2}{|l|}{ConvPoint \cite{convpoint}} & 91.8\%\ & 88.5\%\ & - & - \\
       \hline
        Graph-based & Spectral & DPAM \cite{dpam} &  91.9\%\ & 89.9\%\ & 94.6\%\ & 94.3\%\ \\
        Methods & Methods & RGCNN \cite{rgcnn} &  90.5\%\ & 87.3\%\ & - & - \\
        & & 3DTI-Net \cite{3dtinet} &  91.7\%\ & - & - & - \\
        & & PointGCN \cite{pointgcn} &  89.5\%\ & 86.1\%\ & 91.9\%\ & 91.6\%\ \\
        & & LocalSpecGCN \cite{LSgcn} &  92.1\%\ & - & - & - \\
        \cline{2-7}
        & Spatial & ECC \cite{ecc} &  87.4\%\ & 83.2\%\ & 90.8\%\ & 90.0\%\ \\
        & Methods & KCNet \cite{kcnet} &  91.0\%\ & - & 94.4\%\ & - \\
        & & DGCNN \cite{dgcnn} &  92.2\%\ & 90.2\%\ & - & - \\
        & & LDGCNN \cite{ldgcnn} &  92.9\%\ & 90.3\%\ & - & - \\
        & & Hassani et al. \cite{hassani}  &  89.1\%\ & - & - & - \\
        & & ClusterNet \cite{clusternet} &  87.1\%\ & - & - & - \\ 
        & & Grid-GCN \cite{gridgcn} &  93.1\%\ & 91.3\%\  & 97.5\%\ & 97.4\%\ \\
        & & SAGConv \cite{SAGC} &  93.5\%\ & 91.3\%\  & 98.3\%\ & 97.7\%\ \\
        & & SAMGC \cite{SAMGC} &  93.6\%\ & 91.4\%\  & 98.3\%\ & 97.7\%\ \\
        \cline{3-7}
        & & \textbf{SFAGC} & \textbf{94.0\%\ } & \textbf{91.6\%\ } & \textbf{98.6\%\ } & \textbf{97.8\%\ } \\
        \hline
        \hline
   \end{tabular}}
\end{table*}

\begin{table*}[htb]
\centering  
\footnotesize
\setlength{\abovecaptionskip}{0pt} 
\setlength{\belowcaptionskip}{10pt}
\caption{The hyperparameters of the point cloud segmentation network. $k$ is the value of $k$-NN, $t$ is the number of selected nodes. $\text{co}_{\text{in}}$ is the number of channels of the input coordinates. $\text{f}_{\text{in}}$ is the number of channels of the input node-wise features. $\text{co}_{\text{out}}$ is the number of channels of the output coordinates. $\text{f}_{\text{out}}$ is the number of channels of the output node-wise features. The symbol ‘‘-’’ indicates that the parameters are not available. The input 3D point cloud is $\mathcal{X}=\{\text{x}_i\}_{i=1}^N$.} 
\label{3ps-hp}
\setlength{\tabcolsep}{3mm}{
 \begin{tabular}{r|l|l|l|l|l}
     \hline
     \hline
       Epoch & \multicolumn{5}{|l}{251} \\
       \hline
       Batch size & \multicolumn{5}{|l}{16} \\
       \hline
       Learning rate & \multicolumn{5}{|l}{0.001} \\
       \hline
       Drop out & \multicolumn{5}{|l}{0.4} \\
       \hline
        Phase & Graph convolution layer & Coordinates update & $\textit{k}$ & \multicolumn{2}{|l}{[$\text{co}_\text{in}$,$\text{f}_\text{in}$,$\text{co}_\text{out}$,$\text{f}_\text{out}$]}    \\
       \hline
        \multirow{2}*{Phase1} & SFAGC($co_v=x_v$) & $co^{\prime}_v=\text{MLP}(co_v)$ & 20 & \multicolumn{2}{|l}{[3,3,32,64]} \\
         & SFAGC & $co^{\prime}_v=co_v$ & 20 & \multicolumn{2}{|l}{[32,64,-,64]} \\
       \hline
        \multirow{2}*{Phase2} & SFAGC & $co^{\prime}_v=\text{MLP}(co_v)$ & 20 & \multicolumn{2}{|l}{[3,64,32,64]} \\
         & SFAGC & $co^{\prime}_v=co_v$ & 20 & \multicolumn{2}{|l}{[32,64,-,128]} \\
       \hline
        \multirow{2}*{Phase3} & SFAGC & $co^{\prime}_v=\text{MLP}(co_v)$ & 20 & \multicolumn{2}{|l}{[3,128,128,256]} \\
         & SFAGC & $co^{\prime}_v=co_v$ & 20 & \multicolumn{2}{|l}{[128,256,-,256]} \\
        \hline
        Graph pooling layername & Graph convolution layer & Coordinates update & $\textit{k}$ & $\textit{t}$ & [$\text{co}_\text{in}$,$\text{f}_\text{in}$,$\text{co}_\text{out}$,$\text{f}_\text{out}$]   \\
       \hline
        $\text{GraphPool}_f$ 1 & SFAGC($co_v=x_v$) & $co^{\prime}_v=co_v$ & 36 & 512 & [3,131,3,64] \\
        $\text{GraphPool}_f$ 2 & SFAGC($co_v=x_v$) & $co^{\prime}_v=co_v$ & 64 & 128 & [3,320,3,128] \\
       \hline
        Feature propagation layer name & \multicolumn{5}{|l}{[input channels, output channels]} \\
        \hline
         Feature propagation1 & \multicolumn{5}{|l}{[256+256+128+128,256]} \\
         Feature propagation2 & \multicolumn{5}{|l}{[256+64+64,128]} \\
        \hline
        \hline
   \end{tabular}}
\end{table*}

\begin{table*}[htb]  
\footnotesize
\centering  
\setlength{\abovecaptionskip}{0pt} 
\setlength{\belowcaptionskip}{10pt}
\caption{Comparison results of the 3D point cloud segmentation on the ShapeNet. mIoU indicates the average mIoU of all test instances. The mIoU of each class is also shown. The results are obtained by experimenting with the same hyper-parameters.}  
\label{3ps-results}
 \setlength{\tabcolsep}{0.87mm}{
 \begin{tabular}{l|c|l|l|l|l|l|l|l|l|l|l|l|l|l|l|l|l} 
     \hline
     \hline
     Method & mIoU & air- & bag & cap & car & chair & ear- & guitar & knife & lamp & lap- & motor & mug & pistol & rocket & skate- & table \\
      & & plane & & & & & phone & & & & top & & & & & board & \\
     \hline
     \multirow{2}{*}{PointNet\cite{pn}} & \multirow{2}{*}{83.0} & \multirow{2}{*}{81.5} & \multirow{2}{*}{64.8} & \multirow{2}{*}{77.2} & \multirow{2}{*}{73.5} & \multirow{2}{*}{88.6} & \multirow{2}{*}{68.3} & \multirow{2}{*}{90.4} & \multirow{2}{*}{84.1} & \multirow{2}{*}{80.0} & \multirow{2}{*}{95.1} & \multirow{2}{*}{59.2} & \multirow{2}{*}{91.8} & \multirow{2}{*}{79.7} & \multirow{2}{*}{52.1} & \multirow{2}{*}{72.4} & \multirow{2}{*}{81.6} \\
      & & & & & & &  & & & & & & & & & & \\
     Point- & \multirow{2}{*}{84.7} & \multirow{2}{*}{81.4} & \multirow{2}{*}{73.4} & \multirow{2}{*}{80.6} & \multirow{2}{*}{\textbf{77.8}} & \multirow{2}{*}{89.9} & \multirow{2}{*}{74.5} & \multirow{2}{*}{90.6} & \multirow{2}{*}{85.8} & \multirow{2}{*}{83.5} & \multirow{2}{*}{95.1} & \multirow{2}{*}{\textbf{69.3}} & \multirow{2}{*}{\textbf{94.0}} & \multirow{2}{*}{81.0} & \multirow{2}{*}{58.5} & \multirow{2}{*}{\textbf{74.3}} & \multirow{2}{*}{82.0} \\
     Net++\cite{pn++} & & & & & & &  & & & & & & & & & & \\
     \multirow{2}{*}{DGCNN\cite{dgcnn}} & \multirow{2}{*}{85.0} & \multirow{2}{*}{82.6} & \multirow{2}{*}{79.8} & \multirow{2}{*}{\textbf{85.3}} & \multirow{2}{*}{76.9} & \multirow{2}{*}{90.4} & \multirow{2}{*}{77.1} & \multirow{2}{*}{\textbf{91.0}} & \multirow{2}{*}{86.9} & \multirow{2}{*}{84.0} & \multirow{2}{*}{\textbf{95.6}} & \multirow{2}{*}{61.5} & \multirow{2}{*}{93.0} & \multirow{2}{*}{79.9} & \multirow{2}{*}{58.2} & \multirow{2}{*}{73.7} & \multirow{2}{*}{\textbf{83.2}} \\
      & & & & & & &  & & & & & & & & & & \\
     Point- & \multirow{2}{*}{84.6} & \multirow{2}{*}{82.4} & \multirow{2}{*}{80.3} & \multirow{2}{*}{83.2} & \multirow{2}{*}{76.8} & \multirow{2}{*}{89.9} & \multirow{2}{*}{\textbf{80.6}} & \multirow{2}{*}{90.8} & \multirow{2}{*}{86.7} & \multirow{2}{*}{83.2} & \multirow{2}{*}{95.3} & \multirow{2}{*}{60.1} & \multirow{2}{*}{93.5} & \multirow{2}{*}{$\bf{81.6}$} & \multirow{2}{*}{59.1} & \multirow{2}{*}{73.7} & \multirow{2}{*}{82.3} \\
     ASNL\cite{pnasnl} & & & & & & &  & & & & & & & & & & \\
     \multirow{2}{*}{PCT\cite{PCT}} & \multirow{2}{*}{84.7} & \multirow{2}{*}{83.6} & \multirow{2}{*}{67.6} & \multirow{2}{*}{83.6} & \multirow{2}{*}{75.4} & \multirow{2}{*}{90.1} & \multirow{2}{*}{74.5} & \multirow{2}{*}{90.8} & \multirow{2}{*}{85.8} & \multirow{2}{*}{82.1} & \multirow{2}{*}{95.4} & \multirow{2}{*}{64.0} & \multirow{2}{*}{92.1} & \multirow{2}{*}{81.1} & \multirow{2}{*}{56.2} & \multirow{2}{*}{72.5} & \multirow{2}{*}{\textbf{83.2}} \\
      & & & & & & &  & & & & & & & & & & \\
     \hline
     \multirow{2}{*}{\textbf{SFAGC}} & \multirow{2}{*}{\textbf{85.5}} & \multirow{2}{*}{\textbf{83.7}} & \multirow{2}{*}{\textbf{80.9}} & \multirow{2}{*}{83.5} & \multirow{2}{*}{77.4} & \multirow{2}{*}{\textbf{90.5}} & \multirow{2}{*}{76.2} & \multirow{2}{*}{\textbf{91.0}} & \multirow{2}{*}{\textbf{88.7}} & \multirow{2}{*}{\textbf{84.2}} & \multirow{2}{*}{95.5} & \multirow{2}{*}{67.4} & \multirow{2}{*}{93.7} & \multirow{2}{*}{81.1} & \multirow{2}{*}{\textbf{59.4}} & \multirow{2}{*}{74.1} & \multirow{2}{*}{\textbf{83.2}} \\
      & & & & & & &  & & & & & & & & & & \\
     \hline
     \hline
 \end{tabular}}
\end{table*}

\section{Experiments}
\label{sec:experiments}
In this section, we conduct experiments on 3D point cloud classification and segmentation to validate the proposed GC.

\subsection{3D point cloud classification}
Here, we present the 3D point cloud classification experiment using the 3D point cloud classification network introduced in Section \ref{sec:Implementation}.

\subsubsection{Dataset}
The ModelNet dataset \cite{modelnet} is used in our point cloud classification experiment. 12,308 computer-aided design (CAD) models in 40 categories are included in ModelNet40. In addition, 9,840 CAD models are utilized for training and 2,468 CAD models are used for testing. 4,899 CAD models are included in ModelNet10. They are divided into 3,991 for training and 908 for testing from ten categories. For each CAD models, the CAD mesh faces were evenly sampled with 1,024 points. Initially, all point clouds were normalized to be in a unit sphere.

\subsubsection{Settings and evaluation} 
The settings of hyperparameters are summarized in Table \ref{3dc-hp}. We use the average accuracy of all test instances (OA) and the average accuracy of all shape classes (mAcc) to evaluate the performance of our network.

\subsubsection{Results and discussion}
Table \ref{3dc-results} summarizes the results for point cloud classification. The results of existing methods are taken from their original papers. In terms of both OA and mAcc, our method performs better than the others.  

In the following, we focus on the comparison of our method and graph-based methods. The GCs utilized in DPAM \cite{dpam} are GCN \cite{gcn}. Therefore, it only uses node-wise features of one-hop neighboring nodes. RGCNN \cite{rgcnn}, 3DTI-Net \cite{3dtinet}, PointGCN \cite{pointgcn}, and LocalSpaceGCN \cite{LSgcn} are the spectral methods. Although they can design the global spectral response, they may neglect the local spatial information. In comparison with the direct spatial methods \cite{ecc, kcnet, dgcnn, ldgcnn, hassani, clusternet, gridgcn}, our method can obtain the local structural information of the graph in the feature space, and the information of neighboring nodes can be utilized efficiently using attention-based aggregation. In comparison with the direct spatial methods, i.e., SAGConv \cite{SAGC} and SAMGC \cite{SAMGC}, the proposed method can better aggregate the structural information of the neighboring nodes using the local structure projection aggregation. Furthermore, the information of neighboring nodes can be utilized efficiently using attention-based aggregation. These are possible reasons for the performance improvement of the proposed method.

\begin{table*}[htb]  
\footnotesize
\centering  
\setlength{\abovecaptionskip}{0pt} 
\setlength{\belowcaptionskip}{10pt}
\caption{The components of different AGCs. The symbol "\Checkmark" indicates that the component is contained in AGC. The symbol "\XSolidBrush" indicates that the component is not contained in AGC.}  
\label{different-AGCs}
\setlength{\tabcolsep}{0.3mm}{
 \begin{tabular}{c|c|c|c|c} 
 \hline
 \hline
 \multirow{2}{*}{AGC} & Local structure projection aggregation & Position & \multicolumn{2}{c}{Weighted sum aggregation and update (Section \ref{subsec:Weighted sum aggregation and update})} \\
 \cline{4-5}
   & and fusing (Section \ref{subsec:local structure projection aggregation and fusing})& embedding (Section \ref{subsec:position embedding})& Subtraction attention function & Dot-production attention function \\
 \hline
  SFAGC & \Checkmark & \Checkmark & \Checkmark & \Checkmark \\
  SFAGC-nS & \XSolidBrush & \Checkmark & \Checkmark & \Checkmark \\
  SFAGC-nP & \Checkmark & \XSolidBrush & \Checkmark & \Checkmark \\
  SFAGC-ndot & \Checkmark & \Checkmark & \Checkmark & \XSolidBrush \\
  SFAGC-nsub & \Checkmark & \Checkmark & \XSolidBrush & \Checkmark \\
 \hline
 \hline
 \end{tabular}}
\end{table*}

\begin{figure*}[htb]
\setlength{\abovecaptionskip}{0pt}
\setlength{\belowcaptionskip}{0pt}
\centering
\includegraphics[width=13cm]{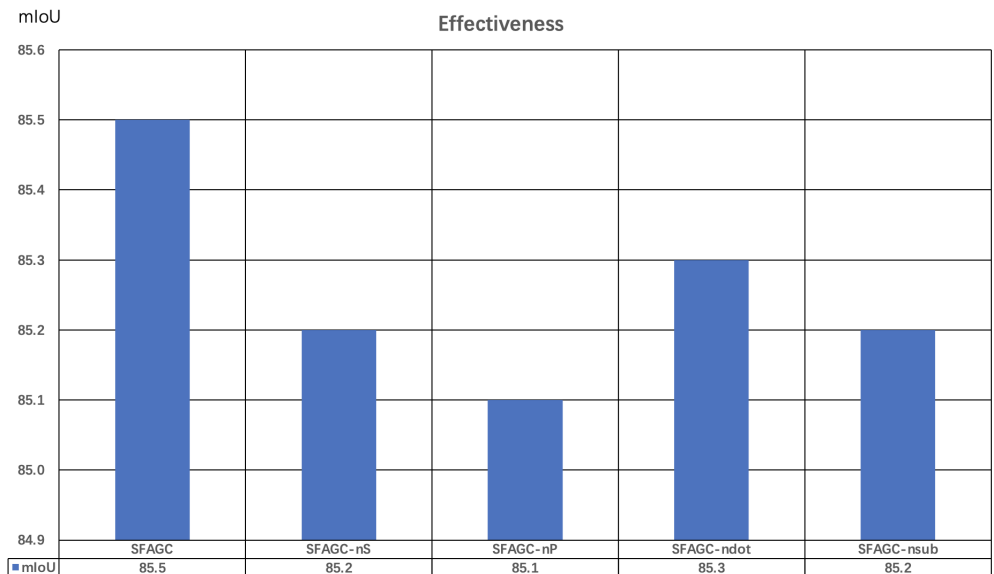}
\centering
\caption{The results of 3D point cloud segmentation experiments confirming the effectiveness of SFAGC.}
\label{fig:effective}
\end{figure*}

\subsection{3D point cloud segmentation}
In this subsection, we also perform a 3D point cloud segmentation experiment.

\subsubsection{Dataset}
The ShapeNet dataset \cite{shapenet} is used in the experiment. It contains 16,846 computer-aided design (CAD) models in 40 categories and each point cloud has 2,048 points. 2,874 point clouds are used for testing, and 13,998 of them are taken for training.

\subsubsection{Settings and evaluation}
Table \ref{3ps-hp} shows the hyperparameter settings. For this experiment, we re-perform existing methods with their corresponding codes available online. The hyperparameters (epoch, batch size, learning rate, and drop out) used in the existing methods experiments are the same with those shown in Table \ref{3ps-hp}. 

We evaluated the average mIoU of all test instances with other neural networks designed specifically for point cloud segmentation.

\subsubsection{Results and discussion} 
Experimental results for point cloud segmentation are summarized in Table \ref{3ps-results}. It is observed that our method has higher mIoU than the existing methods. 

Here, we discuss the possible reasons of the improvement. We first focus on the comparison between our method and DGCNN \cite{dgcnn}. DGCNN corresponds to a graph-based direct spatial method. In contrast to the direct method, our method can utilize the local structural information in the feature space, and it also collects the information of neighboring nodes through attention-based aggregation.

We also compare our method with transformer-based methods PointASNL \cite{pnasnl} and PCT \cite{PCT}. While the transformer basically has a large number of parameters, they are restricted to use only node-wise features and the single dot-production attention function to calculate attention weights. In contrast, SFAGC utilizes the local structural information in the feature space and the multiple-type attention functions.

\subsection{Ablation Study}
We also perform extra 3D point cloud segmentation experiments to validate the effectiveness of the components in the SFAGC.
 
The hyperparameters are the same as those shown in Table \ref{3ps-hp}. Here, we use some AGCs with different settings as follows:
\begin{enumerate}
\item  \textbf{SFAGC}. This is the full SFAGC described in Section \ref{sec:SFPMAGC}.
\item  \textbf{SFAGC-nS}. To validate the effectiveness of the local structure projection aggregation and fusing part proposed in the Section \ref{subsec:local structure projection aggregation and fusing}, we discard this part from the SFAGC.
\item  \textbf{SFAGC-nP}. To confirm the effectiveness of the position embedding proposed in Section \ref{subsec:position embedding}, we bypass the position embedding part from the SFAGC.
\item  \textbf{SFAGC-ndot and SFAGC-nsub}. To validate the effectiveness of the multi-type attention function proposed in Section \ref{subsec:Weighted sum aggregation and update}, SFAGC-ndot is set as the SFAGC without the dot-production attention function, and SFAGC-nsub is set as the SFAGC without the subtraction attention function.
\end{enumerate} 
The architecture of the network and hyperparameters are the same as the previous experiment.

The results are summarized in Fig. \ref{fig:effective}. By comparing SFAGC with SFAGC-nS, use of the local structures in feature space increases 0.3 mIoU. In SFAGC vs. SFAGC-nP, the position-embedding phase increases 0.4 mIoU. In SFAGC vs. SFAGC-ndot, the multi-type attention function increases 0.2 mIoU. In SFAGC vs. SFAGC-nsub, the multi-type attention function increases 0.3 mIoU. The effectiveness of the SFAGC modules was demonstrated in this study.

\section{Conclusion}
In this paper, we propose a new attention-based graph convolution named SFAGC. It can better aggregate the structural information of the neighboring nodes in high-dimensional feature space by using local structure projection aggregation. It also can computes better attention weights by using multi-type attention functions simultaneously. Through experiments on point cloud classification and segmentation, our method outperforms existing methods.

\bibliographystyle{IEEEtranc}

\EOD

\end{document}